\documentclass[11pt,a4paper]{article}

\newcommand{\cmark}{\ding{51}}
\newcommand{\xmark}{\ding{55}}

\newcommand\ti[1]{\textit{#1}}
\newcommand\ts[1]{\textsc{#1}}
\newcommand\tf[1]{\textbf{#1}}

\newcommand\mf[1]{\mathbf{#1}}

\newcommand{\noentity}{$\epsilon$}
\newcommand{\norelation}{$\epsilon$}

\newcommand{\ours}{\textsc{PURE}}

\usepackage[]{naacl2021}
\usepackage{times}
\usepackage{latexsym}
\usepackage{amsmath}
\usepackage{amssymb}
\usepackage{array}
\usepackage{pifont}
\usepackage{microtype}
\usepackage{tabularx}
\usepackage{adjustbox}
\usepackage{enumitem}
\DeclareMathOperator*{\argmax}{arg\,max}
\usepackage{multirow}

\usepackage{booktabs,subcaption,amsfonts,dcolumn}

\usepackage[compact]{titlesec}
\titlespacing{\section}{0pt}{2ex}{1ex}
\titlespacing{\subsection}{0pt}{1ex}{1ex}

\newcolumntype{Y}{>{\centering\arraybackslash}X}

\mathchardef\ordinarycolon\mathcode`\:
\mathcode`\:=\string"8000
\begingroup \catcode`\:=\active
  \gdef:{\mathrel{\mathop\ordinarycolon}}
\endgroup

\title{A Frustratingly Easy Approach for Entity and Relation Extraction}

\author{Zexuan Zhong \quad Danqi Chen \\
Department of Computer Science \\ Princeton University  \\
\texttt{\{zzhong, danqic\}@cs.princeton.edu }
}

\date{}

\begin{document}
\maketitle

\begin{abstract}
%!TEX root = main.tex

% \todo{Add acknowledgements -- you have names listed in the latex.}

% \todo{Add new experiemntal results you had during rebuttal and then reduce the page within 9 pages.}

% \todo{Double check if that 1.7\% - 2.8\% is still correct with the new numbers. Better define a macro.}

% \todo{Add a footnote that our approximation model can't benefit from speed-up during training because of the optimization issue. }

End-to-end relation extraction aims to identify named entities and extract relations between them. Most recent work models these two subtasks jointly, either by casting them in one structured prediction framework, or performing multi-task learning through shared representations. In this work, we present a simple pipelined approach for entity and relation extraction, and establish the new state-of-the-art on standard benchmarks (ACE04, ACE05 and SciERC), obtaining a 1.7\%-2.8\% absolute improvement in relation F1 over previous joint models with the same pre-trained encoders. Our approach essentially builds on two independent encoders and merely uses the entity model to construct the input for the relation model. Through a series of careful examinations, we validate the importance of learning distinct contextual representations for entities and relations, fusing entity information early in the relation model, and incorporating global context.  Finally, we also present an efficient approximation to our approach which requires only one pass of both entity and relation encoders at inference time, achieving an 8-16$\times$ speedup with a slight reduction in accuracy.\footnote{Our code and models are publicly available at \url{https://github.com/princeton-nlp/PURE}.}

\end{abstract}

%!TEX root = main.tex

\section{Introduction}

\begin{figure*}[ht]
  \centering
  \resizebox{2\columnwidth}{!}{%
  \includegraphics{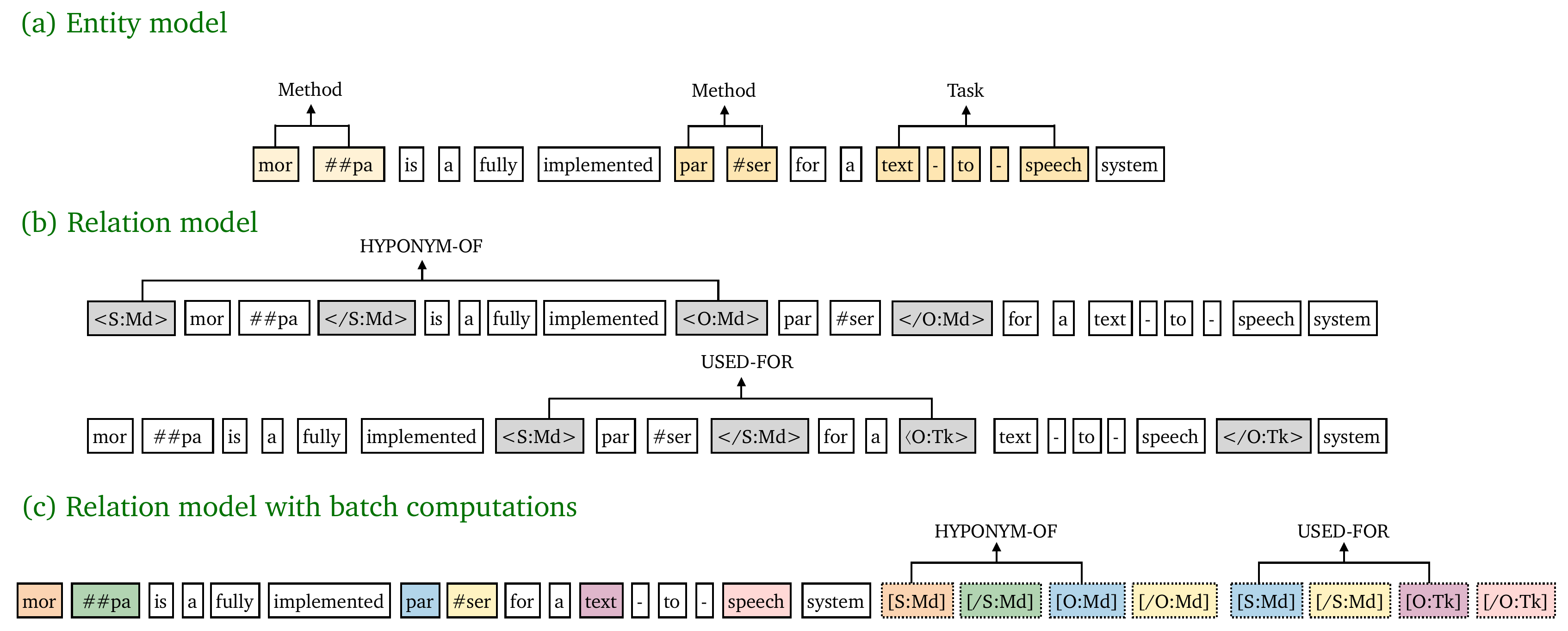}
  }
  % \vspace{-0.5em}
    \caption{An example from the SciERC dataset~\cite{luan2018multi}. Given an input sentence \ti{MORPA is a fully implemented parser for a text-to-speech system}, an end-to-end relation extraction system is expected to extract that \ts{MORPA} and \ts{parser} are entities of type \ts{Method}, \textsc{text-to-speech} is a \ts{Task}, as well as \textsc{MORPA} is a \ti{hyponym of} \textsc{parser} and \textsc{MORPA} is \ti{used for} \textsc{text-to-speech}. (a) Our entity model predicts all the entities at once. (b) Our relation model considers every pair of entities independently by inserting typed entity markers (e.g., \textsc{[S:Md]}:  the subject is a \ts{Method}, \textsc{[O:Tk]}: the object is a \ts{Task}). (c) We also proposed an approximation relation model which supports batch computations. The tokens of the same color share the positional embeddings (see Section~\ref{sec:approx} for more details).}
    \label{fig:mainfig}
\vspace{-1em}
\end{figure*}

Extracting entities and their relations from unstructured text is a fundamental problem in information extraction. This problem can be decomposed into two subtasks: named entity recognition~\cite{sang2003introduction,ratinov2009design} and relation extraction~\cite{zelenko2002kernel,bunescu2005shortest}.
Early work employed a pipelined approach, training one model to extract entities~\cite{florian2004statistical,florian2006factorizing}, and another model to classify relations between them~\cite{zhou2005exploring,kambhatla2004combining,chan2011exploiting}.
More recently, however, end-to-end evaluations have been dominated  by systems that model these two tasks jointly~\cite{li2014incremental,miwa2016end,katiyar2017going,zhang2017end,li2019entity,luan2018multi,luan2019general,wadden2019entity,lin2020joint,wang2020two}. There has been a long held belief that joint models can better capture the interactions between entities and relations and help mitigate error propagation issues.

In this work, we re-examine this problem and present a simple approach which learns \ti{two} encoders built on top of deep pre-trained language models~\cite{devlin2019bert,beltagy2019scibert,lan2020albert}. The two models --- which we refer them as to the \ti{entity model} and \ti{relation model} throughout the paper --- are trained independently and the relation model only relies on the entity model to provide input features. Our entity model builds on span-level representations and our relation model builds on contextual representations specific to \ti{a given pair of spans}. Despite its simplicity, we find this pipelined approach to be extremely effective: using the same pre-trained encoders, our model outperforms all previous joint models on three standard benchmarks: ACE04, ACE05 and SciERC, advancing the previous state-of-the-art by 1.7\%--2.8\% absolute in relation F1.

To better understand the effectiveness of this approach, we carry out a series of careful analyses. We observe that, (1) the contextual representations for the entity and relation models essentially capture distinct information, so sharing their representations hurts performance; (2) it is crucial to fuse the entity information (both boundary and type) at the \ti{input layer} of the relation model; (3) leveraging cross-sentence information is useful in both tasks.  Hence, we expect that this simple model will serve as a very strong baseline in end-to-end relation extraction and make us rethink the value of joint modeling of entities and relations.
% (4) stronger pre-trained language models can bring further gains.

Finally, one possible shortcoming of our approach is that we need to run our relation model once for every pair of entities. To alleviate this issue, we present a novel and efficient alternative by approximating and batching the computations for different groups of entity pairs at inference time. This approximation achieves an 8-16$\times$ speedup with only a slight reduction in accuracy (e.g., $1.0\%$ F1 drop on ACE05), which makes our model fast and accurate to use in practice. Our final system is called {\ours} (the \tf{P}rinceton \tf{U}niversity \tf{R}elation \tf{E}xtraction system) and we make our code and models publicly available for the research community.

We summarize our contributions as follows:
\begin{itemize}[noitemsep,topsep=0pt]
    \item We present a simple and effective approach for end-to-end relation extraction, which learns two independent encoders for entity recognition and relation extraction. Our model establishes the new state-of-the-art on three standard benchmarks and surpasses all previous joint models.
    \item We conduct careful analyses to understand why our approach performs so well and how different factors impact the final performance. We conclude that it is more effective to learn distinct contextual representations for entities and relations than to learn them jointly.
    \item To speed up the inference time of our model, we also propose a novel efficient approximation, which achieves a large runtime improvement with only a small accuracy drop.
\end{itemize}

\label{sec:intro}

%!TEX root = main.tex

% \input{tables/comparison_previous_approaches}

\section{Related Work}
\label{sec:related}

Traditionally, extracting relations between entities in text has been studied as two separate tasks: named entity recognition and relation extraction. In the last several years, there has been a surge of interest in developing models for joint extraction of entities and relations~\cite{li2014incremental,miwa2014modeling,miwa2016end}.  We group existing joint models into two categories: \ti{structured prediction} and \ti{multi-task learning}:

% and discuss their differences below. At the end, we briefly review neural model architectures used in relation extraciton.

% \vspace{-0.4em} 
\paragraph{Structured prediction} Structured prediction approaches cast the two tasks into one unified framework, although it can be formulated in various ways. \newcite{li2014incremental} propose an action-based system which identifies new entities as well as links to previous entities, \newcite{zhang2017end,wang2020two} adopt a table-filling approach proposed in \cite{miwa2014modeling};
\newcite{katiyar2017going} and \newcite{zheng2017joint} employ sequence tagging-based approaches; \newcite{sun2019joint} and \newcite{fu2019graphrel} propose graph-based approaches to jointly predict entity and relation types;
and, \newcite{li2019entity} convert the task into a multi-turn question answering problem. All of these approaches need to tackle a global optimization problem and perform joint decoding at inference time, using beam search or reinforcement learning.

% \danqi{TODO: add \cite{wang2020two}}

% \vspace{-0.4em}
 \paragraph{Multi-task learning} This family of models essentially builds two separate models for entity recognition and relation extraction and optimizes them together through parameter sharing. \newcite{miwa2016end} propose to use a sequence tagging model for entity prediction and a tree-based LSTM model for relation extraction.
The two models share one LSTM layer for contextualized word representations and they find sharing parameters improves performance (slightly) for both models. The approach of \newcite{bekoulis2018adversarial} is similar except that they model relation classification as a multi-label head selection problem. Note that these approaches still perform pipelined decoding: entities are first extracted and the relation model is applied on the predicted entities.

% use a sequence tagging model for entity prediction and a form the relation classification as a multi-label prediction selection, where two parts shared one LSTM.
% The entity model was first pre-trained and then both models were optimized jointly.

The closest work to ours is DYGIE and DYGIE++~\cite{luan2019general,wadden2019entity}, which builds on recent span-based models for coreference resolution~\cite{lee2017end} and semantic role labeling~\cite{he2018jointly}. The key idea of their approaches is to learn shared span representations between the two tasks and update span representations through dynamic graph propagation layers.
% It is later improved by incorporating relation and coreference propagation layers to update span representations~\cite{luan2019general} and replacing LSTM encodings with BERT-based representations~\cite{wadden2019entity}.
A more recent work~\newcite{lin2020joint} further extends DYGIE++ by incorporating global features based on cross-substask and cross-instance constraints.\footnote{This is an orthogonal contribution to ours and we will explore it for future work.} Our approach is much simpler and we will detail the differences in Section~\ref{sec:our-model} and explain why our model performs better.

\section{Method}
\label{sec:method}

In this section, we first formally define the problem of end-to-end relation extraction in Section~\ref{sec:problem-definition} and then detail our approach in Section~\ref{sec:our-model}.  Finally, we present our approximation solution in Section~\ref{sec:method-approx}, which considerably improves the efficiency of our approach during inference.

\subsection{Problem Definition}
\label{sec:problem-definition}
% End-to-end relation extraction problem consists of two sub-tasks: named entity recognition and relation extraction.
The input of the problem is a sentence $X$ consisting of $n$ tokens $x_1, x_2, \dots, x_n$. Let $S=\{s_1, s_2, \dots, s_m\}$ be all the possible spans in $X$ of up to length $L$ and $\ts{START}(i)$ and $\ts{END}(i)$ denote start and end indices of $s_i$. Optionally, we can incorporate cross-sentence context to build better contextual representations (Section \ref{sec:our-model}). The problem can be decomposed into two sub-tasks:
\paragraph{Named entity recognition} Let $\mathcal{E}$ denote a set of pre-defined entity types. The named entity recognition task is, for each span $s_i \in S$, to predict an entity type $y_e(s_i) \in \mathcal{E}$ or $y_e(s_i) = \epsilon$ representing span $s_i$ is not an entity. The output of the task is $Y_e = \{(s_i, e): s_i \in S, e \in \mathcal{E}\}$.
\paragraph{Relation extraction} Let $\mathcal{R}$ denote a set of pre-defined relation types. The task is, for every pair of spans $s_i \in S, s_j\in S$, to predict a relation type $y_r(s_i, s_j) \in \mathcal{R}$, or there is no relation between them: $y_r(s_i, s_j) = \epsilon$. The output of the task is $Y_r = \{(s_i, s_j, r): s_i, s_j \in S, r \in \mathcal{R}\}$.

% We aim to build a model which takes $X$ as input and outputs $Y_e$ and $Y_r$ at the same time. During evaluation, $Y_e$ and $Y_r$ are compared against the ground truth $Y^*_e$ and $Y^*_r$ and entity and relation F1 will be reported respectively.\footnote{The strict evaluation of relation F1 also takes the entity type into account.  See Section~\ref{sec:exp-setup} for more details.}

\subsection{Our Approach}
\label{sec:our-model}
% In the following, we will describe our full model which consists of an entity model and a relation model.
As shown in Figure~\ref{fig:mainfig}, our approach consists of an entity model and a relation model.
The entity model first takes the input sentence and predicts an entity type (or \noentity) for each single span. We then process every pair of candidate entities independently in the relation model by inserting extra marker tokens to highlight the subject and object and their types. We will detail each component below, and finally summarize the differences between our approach and DYGIE++~\cite{wadden2019entity}.

\paragraph{Entity model} Our entity model is a standard span-based model following prior work~\cite{lee2017end,luan2018multi,luan2019general,wadden2019entity}. We first use a pre-trained language model (e.g., BERT) to obtain contextualized representations $\mf{x}_t$ for each input token $x_t$. Given a span $s_i \in S$, the span representation $\mathbf{h}_e(s_i)$ is defined as:
\vspace{-0.3em}
\begin{align*}
  \mathbf{h}_e(s_i) = [\mathbf{x}_{\textsc{START}(i)}; \mathbf{x}_{\textsc{END}(i)}; \phi(s_i)],
\end{align*}
where $\phi(s_i) \in \mathbb{R}^{d_F}$ represents the learned embeddings of span width features.
The span representation $\mathbf{h}_e(s_i)$ is then fed into a feedforward network to predict the probability distribution of the entity type $e \in \mathcal{E} \cup \{\epsilon\}$: $P_e(e \mid s_i)$.
% The span representation $\mathbf{h}_e(s_i)$ is then used to predict entity types $e \in \mathcal{E} \cup \{\epsilon\}$:
% \vspace{-0.3em}
% \begin{align*}
%     P_e(e \mid s_i) = \mathrm{softmax}(\mathbf{W}_e \text{FFNN}(\mathbf{h}_e(s_i)).
% \end{align*}
% We follow~\newcite{wadden2019entity} and use a 2-layer feedforward neural network with ReLU activations.

\paragraph{Relation model} The relation model aims to take a pair of spans $s_i, s_j$ (a subject and an object) as input and predicts a relation type or \norelation. Previous approaches~\cite{luan2018multi,luan2019general,wadden2019entity} re-use the span representations $\mf{h}_e(s_i), \mf{h}_e(s_j)$ to predict the relationship between $s_i$ and $s_j$. We hypothesize that these representations only capture contextual information around each individual entity and might fail to capture the dependencies between the pair of spans. We also argue that sharing the contextual representations between \ti{different} pairs of spans may be suboptimal. For instance, the words \ti{is a} in ~Figure~\ref{fig:mainfig} are crucial in understanding the relationship between \ts{MORPA} and \ts{parser} but not for \ts{MORPA} and \ts{text-to-speech}.

Our relation model instead processes each pair of spans independently and inserts typed markers at the input layer to highlight the subject and object and their types. Specifically, given an input sentence $X$ and a pair of subject-object spans $s_i, s_j$, where $s_i$, $s_j$ have a type of $e_i, e_j \in \mathcal{E} \cup \{\epsilon\}$ respectively.
We define text markers as $\langle \textsc{S:}e_i\rangle$, $\langle \textsc{/S:}e_i\rangle$, $\langle \textsc{O:}e_j\rangle$, and $\langle \textsc{/O:}e_j\rangle$,
% \texttt{[/o:$e_j$]},
and insert them into the input sentence before and after the subject and object spans (Figure~\ref{fig:mainfig} (b)).\footnote{Our final model indeed only considers $e_i, e_j \neq \epsilon$. We have explored strategies using spans which are predicted as $\epsilon$ for the relation model but didn't find improvement. See Section~\ref{sec:error-prop} for more discussion.}
Let $\widehat{X}$ denote this modified sequence with text markers inserted:
% into the input document to identify the spans:
\vspace{-0.3em}
\begin{align*}
    &\widehat{X} = \dots \langle \textsc{S:}e_i\rangle, x_{\textsc{START}(i)}, \dots, x_{\textsc{END}(i)}, \langle \textsc{/S:}e_i\rangle, \\
    &\dots \langle \textsc{O:}e_j\rangle, x_{\textsc{START}(j)}, \dots, x_{\textsc{END}(j)}, \langle \textsc{/O:}e_j\rangle, \dots.
\end{align*}
% &\widehat{X} = \dots [\text{sub:}e_i], x_{\textsc{START}(i)}, \dots, x_{\textsc{END}(i)}, [/\text{sub:}e_i], \\
% &\dots [\text{obj:}e_j], x_{\textsc{START}(j)}, \dots, x_{\textsc{END}(j)}, [/\text{obj:}e_j], \dots

We apply a \ti{second} pre-trained encoder on $\widehat{X}$ and denote the output representations by $\mathbf{\widehat{x}}_t$. We concatenate the output representations  of two start positions and obtain the span-pair representation:
\begin{align*}
    \mathbf{h}_r(s_i, s_j)=[\mathbf{\widehat{x}}_{\widehat{\textsc{START}}(i)}; \mathbf{\widehat{x}}_{\widehat{\textsc{START}}(j)}],
\end{align*}
where $\widehat{\textsc{START}(i)}$ and $\widehat{\textsc{START}(j)}$ are the indices of  $\langle \textsc{S:}e_i\rangle$ and $\langle \textsc{O:}e_j\rangle$ in $\widehat{X}$. Finally, the representation $\mathbf{h}_r(s_i, s_j)$ will be fed into a feedforward network to predict the probability distribution of the relation type $r \in \mathcal{R} \cup \{\epsilon\}$: $P_r(r|s_i, s_j)$.
% used to predict the relation type $r \in \mathcal{R} \cup \{\epsilon\}$:
% \begin{align*}
%     P_r(r|s_i, s_j) = \mathrm{softmax}(\mathbf{W}_r \mathbf{h}_r(s_i, s_j)),
% \end{align*}

% . Let  denote the output representations of BERT based on the modified input $\widehat{X}$.
% \danqi{This $\textsc{START}(i)-1$ is inaccurate as you might insert 4 markers in total. Change a notation maybe?}
% The representation  is fed into another 2-layer FFNN feedforward layer $\text{FFNN}_r$ followed by a softmax layer over $\mathcal{R} \cup \{\epsilon\}$:

% We also use a 2-layer FFNN \danqi{non-linearity/LN?} as.
% (2-layer, with GeLU activations and layer normalization, following~\newcite{joshi2020spanbert})

% \danqi{Can you add the dimensions of $\mathbf{W}_r$ and maybe $\text{FFNN}_r$? Same for the entity model.}

This idea of using additional markers to highlight the subject and object is not entirely new as it  has been studied recently in relation classification ~\cite{zhang2019ernie,soares2019matching,peters2019knowledge}. However, most relation classification tasks (e.g., TACRED~\cite{zhang2017tacred}) only focus on a given pair of subject and object in an input sentence and its effectiveness has not been evaluated in the end-to-end setting in which we need to classify the relationships between multiple entity mentions. We observed a large improvement in our experiments (Section~\ref{sec:input-features})  and this strengthens our hypothesis that modeling the relationship between different entity pairs in one sentence require different contextual representations. Furthermore, \newcite{zhang2019ernie,soares2019matching} only consider untyped markers (e.g., $\langle\textsc{S}\rangle$, $\langle\textsc{/S}\rangle$) and previous end-to-end models (e.g., \cite{wadden2019entity}) only inject the entity type information into the relation model through auxiliary losses.   We find that injecting type information at the input layer is very helpful in distinguishing entity types --- for example, whether ``Disney'' refers to a \ti{person} or an \ti{organization}--- before trying to understand the relations.

% Using \ti{typed} entity markers hasn't been explored before.

\paragraph{Cross-sentence context}
Cross-sentence information can be used to help predict entity types and relations, especially for pronominal mentions.
\newcite{luan2019general,wadden2019entity} employ a propagation mechanism to incorporate cross-sentence context. \newcite{wadden2019entity} also add a 3-sentence context window which is shown to improve performance.
We also evaluate the importance of leveraging cross-sentence context in our approach. As we expect that pre-trained language models to be able to capture long-range dependencies, we simply incorporate cross-sentence context by extending the sentence to a fixed window size $W$ for both the  entity and relation model. Specifically, given an input sentence with $n$ words, we augment the input with $(W-n) / 2$ words from the left context and right context respectively.
% In our default settings, we use $W=300$ for entity models and use $W=100$ for relation models.

\paragraph{Training \& inference} For both entity model and relation model, we fine-tune the two pre-trained language models using task-specific losses. We use cross-entropy loss for both models:
\begin{align*}
    \mathcal{L}_e &= -\sum_{s_i \in S} \log P_e(e_i^* | s_i) \\
    \mathcal{L}_r &= -\sum_{s_i, s_j \in S_G, s_i \neq s_j} \log P_r(r_{i,j}^* \mid s_i, s_j),
\end{align*}
where $e_i^*$ represents the gold entity type of $s_i$ and $r_{i,j}^*$ represents the gold relation type of span pair $s_i, s_j$ in the training data. For training the relation model, we only consider the gold entities $S_G \subset S$ in the training set and use the gold entity labels as the input of the relation model. We considered training on predicted entities as well as all spans $S$ (with pruning), but none of them led to meaningful improvements compared to this simple pipelined training (see more discussion in Section~\ref{sec:error-prop}). During inference, we first predict the entities by taking $y_e(s_i) = \argmax_{e\in \mathcal{E} \cup \{\epsilon\}}P_e(e | s_i)$. Denote $S_{\text{pred}} = \{s_i: y_e(s_i) \neq \epsilon\}$, we enumerate all the spans $s_i, s_j \in S_{\text{pred}}$ and use $y_e(s_i), y_e(s_j)$ to construct the input for the relation model $P_r(r \mid s_i, s_j)$.

% \vspace{-0.3em}
\paragraph{Differences from DYGIE++} Our approach differs from DYGIE++~\cite{luan2019general,wadden2019entity} in the following ways: (1) We use separate encoders for the entity and relation models, without any multi-task learning. The predicted entity types are used directly to construct the input for the relation model. (2) The contextual representations in the relation model are specific to each pair of spans by using the text markers. (3) We only incorporate cross-sentence information by extending the input with additional context (as they did) and we do not employ any graph propagation layers and beam search.\footnote{They also incorporated coreferences and event prediction in their framework. We focus on entity and relation extraction in this paper and we leave these extensions to future work.} As a result, our model is much simpler. As we will show in the experiments (Section~\ref{sec:exp}), it also achieves large gains in all the benchmarks, using the same pre-trained encoders.

\subsection{Efficient Batch Computations}
\label{sec:method-approx}
% Despite the simplicity and effectiveness of our approach (which we will demonstrate in our experiments), one possible shortcoming is that we need to run our relation model once for every pair of entities.
One possible shortcoming of our approach is that we need to run our relation model once for every pair of entities.
To alleviate this issue, we propose a novel and efficient alternative to our relation model. The key problem is that we would like to re-use computations for different pairs of spans in the same sentence. This is impossible in our original model because we must insert the entity markers for each pair of spans independently. To this end, we propose an approximation model by making two major changes to the original relation model.  First, instead of directly inserting entity markers into the original sentence, we tie the position embeddings of the markers with the start and end tokens of the corresponding span:
\begin{align*}
    & \textsc{p}(\langle \textsc{S:}e_i\rangle),\textsc{p}(\langle \textsc{/S:}e_i\rangle) := \textsc{p}(x_{\textsc{START}(i)}),\textsc{p}(x_{\textsc{END}(i)}) \\
    % & \textsc{pos}(\langle \textsc{/S:}e_i\rangle) = \textsc{pos}(x_{\textsc{END}(i)}) \\
    & \textsc{p}(\langle \textsc{O:}e_j\rangle),\textsc{p}(\langle \textsc{/O:}e_j\rangle) := \textsc{p}(x_{\textsc{START}(j)}),\textsc{p}(x_{\textsc{END}(j)}),
    % & \textsc{pos}(\langle \textsc{/O:}e_j\rangle) = \textsc{pos}(x_{\textsc{END}(j)}),
\end{align*}
where $\textsc{p}(\cdot)$ denotes the position id of a token. As the example shown in Figure~\ref{fig:mainfig}, if we want to classify the relationship between \ts{MORPA} and \ts{parser}, the first entity marker \ts{$\langle S$$:\ts{Method}\rangle$} will share the position embedding with the token \ts{mor}. By doing this, the position embeddings of the original tokens will not be changed.

Second, we add a constraint to the attention layers. We enforce the text tokens to only attend to text tokens and not attend to the marker tokens while an entity marker token can attend to all the text tokens and all the 4 marker tokens  associated with the same span pair. These two modifications allow us to re-use the computations of all text tokens, because the representations of text tokens are independent of the entity marker tokens.
Thus, we can batch multiple pairs of spans from the same sentence in one run of the relation model. In practice, we add all marker tokens to the end of the sentence to form an input that batches a set of span pairs (Figure~\ref{fig:mainfig}(c)). This leads to a large speedup at inference time and only a small drop in performance (Section~\ref{sec:approx}).

% \footnote{This may also speedup the training process. However, we observe that the training becomes not stable after batching multiple pairs in one input. We thus train the model without batching pairs and leave addressing the instability as future work.}

%!TEX root = main.tex

%!TEX root = ../main.tex

\begin{table*}[ht]
\begin{center}
\scalebox{0.85}{
\begin{tabular}{l r ccc ccc ccc}
    \toprule
    \multirow{2}{*}{{\tf{Model}}} & \multirow{2}{*}{\tf{Encoder}} & \multicolumn{3}{c}{\tf{ACE05}} & \multicolumn{3}{c}{\tf{ACE04}} & \multicolumn{3}{c}{\tf{SciERC}} \\
    & & {Ent} & {Rel} & {Rel+} & {Ent} & {Rel} & {Rel+} & {Ent} & {Rel} & {Rel+} \\
    \midrule
    \cite{li2014incremental} & - & 80.8 & 52.1 & 49.5 & 79.7 & 48.3 & 45.3 & - & - & - \\
    \cite{miwa2016end} & L & 83.4 & - & 55.6 & 81.8 & - & 48.4 & - & - & - \\
    \cite{katiyar2017going} & L & 82.6 & 55.9 & 53.6 & 79.6 & 49.3 & 45.7 & - & - & - \\
    \cite{zhang2017end} & L & 83.6 & - & 57.5 & - & - & - & - & - & - \\
    \cite{luan2018multi}$^\clubsuit$$^\dagger$ & L+E & - & - & - & - & - & - & 64.2 & 39.3 & - \\
    \cite{luan2019general}$^\clubsuit$$^\dagger$ & L+E & 88.4 & 63.2 & - & 87.4 & 59.7 & - & 65.2 & 41.6 & - \\
    \cite{li2019entity} & Bl & 84.8 & - & 60.2 & 83.6 & - & 49.4 & - & - & - \\
    \cite{dixit2019span} & L+E & 86.0 & - & 62.8 & - & - & - & - & - & - \\
    \cite{wadden2019entity}$^\clubsuit$$^\dagger$ & Bb & 88.6 & 63.4 & - & - & - & - & - & - & \\
    \cite{wadden2019entity}$^\clubsuit$$^\dagger$ & SciB & - & - & - & - & - & - & 67.5 & 48.4 & \\
    \cite{lin2020joint} & Bl & 88.8 & 67.5 & - & - & - & - & - & - & - \\
    \cite{wang2020two} & ALB & 89.5 & 67.6 & 64.3 & 88.6 & 63.3 & 59.6 & - & - & - \\
    \midrule

    \multirow{3}{*}{{\ours} (ours): single-sentence}  & Bb & 88.7 & 66.7 & 63.9 & 88.1 & 62.8 & 58.3 & - & - & - \\
    & SciB & - & - & - & - & - & - & 66.6 & 48.2 & 35.6 \\
    & ALB & 89.7 & 69.0 & 65.6 & 88.8 & 64.7 & 60.2 & - & - & - \\
    \midrule
    \multirow{3}{*}{{\ours} (ours): cross-sentence$^\clubsuit$} & Bb & 90.1 & 67.7 & 64.8 & 89.2 & 63.9 & 60.1 & - & - & - \\
    & SciB & - & - & - & - & - & - & \tf{68.9} & \tf{50.1}  & \tf{36.8}  \\
    & ALB & \tf{90.9} & \tf{69.4} & \tf{67.0} & \tf{90.3} & \tf{66.1} & \tf{62.2} & - & - & - \\

    % \midrule
    % \multirow{7}{*}{{\rotatebox[origin=c]{90}{ACE04}}} & \cite{li2014incremental} & 79.7 & 48.3 & 45.3 \\
    % & \cite{miwa2016end} & 81.8 & - & 48.4 \\
    % & \cite{katiyar2017going} & 79.6 & 49.3 & 45.7 \\
    % & \cite{li2019entity} & 83.6 & - & 49.4 \\
    % & \cite{luan2019general}$\dagger$ & 87.4 & 59.7 & - \\
    % & \cite{wang2020two} & 88.6 & 63.3 & 59.6 \\
    % & Ours & & & \\

    % \midrule
    % \multirow{9}{*}{{\rotatebox[origin=c]{90}{ACE05}}} & \cite{li2014incremental} & 80.8 & 52.1 & 49.5 \\
    % & \cite{miwa2016end} & 83.4 & - & 55.6 \\
    % & \cite{katiyar2017going} & 82.6 & 55.9 & 53.6 \\
    % & \cite{zhang2017end} & 83.6 & - & 57.5 \\
    % & \cite{luan2019general}$\dagger$ & 88.4 & 63.2 & - \\
    % & \cite{li2019entity} & 84.8 & - & 60.2 \\
    % & \cite{wadden2019entity}$\dagger$ & 88.6 & 63.4 & - \\
    % & \cite{lin2020joint} & 88.8 & 67.5 & - \\
    % & \cite{wang2020two} & 89.5 & 67.6 & 64.3 \\
    % & Ours & \tf{90.9} & & \\
    % % Ours & \tf{90.1} & \tf{90.5} & \tf{90.3} & \tf{67.5} & \tf{66.7} & \tf{67.1}  \\

    % \midrule
    % \multirow{4}{*}{{\rotatebox[origin=c]{90}{SciERC}}} & \cite{luan2018multi} & 64.2 & 39.3 & - \\
    % & \cite{luan2019general} & 65.2 & 41.6 & - \\
    % & \cite{wadden2019entity} & 67.5 & 48.4 & - \\
    % & Ours & \tf{68.9} & \tf{50.1} & \\
    \bottomrule
\end{tabular}
}
\caption{Test F1 scores on ACE04, ACE05, and SciERC. We evaluate our approach in two settings: \ti{single-sentence} and \ti{cross-sentence} depending on whether cross-sentence context is used or not. $^\clubsuit$: These models leverage cross-sentence information. $^\dagger$: These models are trained with additional data (e.g., coreference).  The encoders used in different models: L = LSTM, L+E = LSTM + ELMo, Bb = BERT-base, Bl = BERT-large, SciB = SciBERT (size as BERT-base), ALB = ALBERT-xxlarge-v1.
\ti{Rel} denotes the \ti{boundaries} evaluation (the entity boundaries must be correct) and \ti{Rel+} denotes the \ti{strict} evaluation (both the entity boundaries and types must be correct).
% \zexuan{cross-sentence ALB numbers have been updated (previous numbers are only based on 1 run).
}
% \vspace{-0.5em}
\label{tab:main-results}
\end{center}
\end{table*}

\section{Experiments}
\label{sec:exp}

\subsection{Setup}
\label{sec:exp-setup}

%!TEX root = ../main.tex

\begin{table}[t]
\centering
\scalebox{0.93}{
        \begin{tabularx}{\columnwidth}{l*{6}{Y}}
        \toprule
        \tf{Dataset} & $|\mathcal{E}|$ & $|\mathcal{R}|$ & \multicolumn{3}{c}{\# \tf{Sentences}} \\
        & & & Train & Dev & Test \\
        \midrule
        ACE05 & 7 & 6 & 10,051 & 2,424 & 2,050 \\
        ACE04 & 7 & 6 & \multicolumn{3}{c}{$8,683$ (5-fold)}\\
        SciERC & 6 & 7 & 1,861 & 275 & 551 \\
        \bottomrule
        \end{tabularx}}
     \caption{The statistics of the datasets. We use ACE04, ACE05, and SciERC for evaluating end-to-end relation extraction.}
     \label{tab:data_stats}
\end{table}

\paragraph{Datasets} We evaluate our approach on three popular end-to-end relation extraction datasets: ACE05\footnote{\url{catalog.ldc.upenn.edu/LDC2006T06}}, ACE04\footnote{\url{catalog.ldc.upenn.edu/LDC2005T09}}, and SciERC~\cite{luan2018multi}.
Table~\ref{tab:data_stats} shows the data statistics of each dataset.
The ACE05 and ACE04 datasets are collected from a variety of domains, such as newswire and online forums. The SciERC dataset is collected from 500 AI paper abstracts and defines scientific terms and relations specially for scientific knowledge graph construction. We follow previous work and use the same preprocessing procedure and splits for all datasets. See Appendix~\ref{app:datasets} for more details.

% We follow \newcite{luan2019general}'s preprocessing steps and split ACE04 into 5 folds, and split ACE05 and SciERC into train, development, and test sets.
% The detailed data statistics are given in Table~\ref{tab:data_stats}. \danqi{You need 1-2 sentences to describe what is ACE04, ACE05, especially SciERC is about scientific domain.... }

% \input{tables/data_stat.tex}

\paragraph{Evaluation metrics}
% We use the same evaluation criteria as in previous work~\cite{wadden2019entity}.
We follow the standard evaluation protocol and use micro F1 measure as the evaluation metric. For named entity recognition, a predicted entity is considered as a correct prediction if its span boundaries and the predicted entity type are both correct.
For relation extraction, we adopt two evaluation metrics: (1) \ti{boundaries} evaluation (Rel): a predicted relation is considered as a correct prediction if the boundaries of two spans are correct and the predicted relation type is correct; (2)
\ti{strict} evaluation (Rel+): in addition to what is required in the \ti{boundaries} evaluation, predicted entity types also must be correct. More discussion of the evaluation settings can be found in \newcite{bekoulis2018adversarial,taille2020sincere}.

% a predicted relation is considered as a correct prediction if the entity types and boundaries of two spans are correct, as well as the predicted relation type is correct.

\paragraph{Implementation details}
 We use \ti{bert-base-uncased}~\cite{devlin2019bert} and \ti{albert-xxlarge-v1}~\cite{lan2020albert} as the base encoders for ACE04 and ACE05, for a fair comparison with previous work and an investigation of small vs large pre-trained models.\footnote{ As detailed in Table~\ref{tab:main-results}, some previous work used BERT-large models. We are not able to do a comprehensive study of all the pre-trained models and our BERT-base results are generally higher than most published results using larger models.} We also use \ti{scibert-scivocab-uncased}~\cite{beltagy2019scibert} as the base encoder for SciERC, as this in-domain pre-trained model is shown to be more effective than BERT~\cite{wadden2019entity}.
 We use a context window size of $W=300$ for the entity model and $W=100$ for the relation model in our default setting using cross-sentence context\footnote{We use a context window size $W=100$ for the ALBERT entity models to reduce GPU memory usage.} and the effect of different context sizes is provided in Section~\ref{sec:context}. We consider spans up to $L = 8$ words. For all the experiments, we report the averaged F1 scores of 5 runs.
More implementation details can be found in Appendix~\ref{app:impl-details}.

\subsection{Main Results}

% \paragraph{End-to-end relation extraction}
Table~\ref{tab:main-results} compares our approach {\ours} to all the previous results.
We report the F1 scores in both single-sentence and cross-sentence settings. As is shown, our single-sentence models achieve strong performance and incorporating cross-sentence context further improves the results considerably.
Our BERT-base (or SciBERT) models achieve similar or better results compared to all the previous work including models built on top of larger pre-trained LMs, and our results are further improved by using a larger encoder ALBERT.
% using a larger encoder i.e., ALBERT, further improves the performance.

% We use BERT-base and ALBERT-xxlarge as the base encoder for our models.

For entity recognition, our best model achieves an absolute F1 improvement of $+1.4\%$, $+1.7\%$, $+1.4\%$ on ACE05, ACE04, and SciERC respectively.
This shows that cross-sentence information is useful for the entity model and pre-trained Transformer encoders are able to capture long-range dependencies from a large context. For relation extraction, our approach outperforms the best previous methods by an absolute F1 of $+1.8\%$, $+2.8\%$, $+1.7\%$ on ACE05, ACE04, and SciERC respectively.
We also obtained a $4.3\%$ higher relation F1 on ACE05 compared to DYGIE++~\cite{wadden2019entity} using the same BERT-base pre-trained model.
Compared to the previous best approaches using either global features~\cite{lin2020joint} or complex neural models (e.g., MT-RNNs)~\cite{wang2020two}, our approach is much simpler and achieves large improvements on all the datasets.
% It is worth noting that the previous best results come from more complicated approaches, either relying on global features~\cite{lin2020joint}, or sophisticated architecutures, e.g., MT-RNNs~\cite{wang2020two}.
Such improvements demonstrate the effectiveness of learning representations for entities and relations of different entity pairs, as well as early fusion of entity information in the relation model. We also noticed that compared to the previous state-of-the-art model~\cite{wang2020two} based on ALBERT, our model achieves a similar entity F1 (89.5 vs 89.7) but a substantially better relation F1 (67.6 vs 69.0) without using context. This clearly demonstrates the superiority of our relation model.
Finally, we also compare our model to a joint model (similar to DYGIE++) of different data sizes to test the generality of our results. As shown in Appendix~\ref{app:perf-data-sizes}, our findings are robust to data sizes.

% All these large improvements demonstrate the effectiveness of learning distinct representations for entities and relations of different entity pairs, as well as fusing entity information at the input layer of the relation model.
%
% using markers to provide span boundaries and type features at the input layer of the relation model, so that the model is able to learn better contextualized representations specific to the given span pair.

%!TEX root = ../main.tex

\begin{table}[!t]
\begin{center}
\scalebox{0.9}{
\begin{tabular}{l cc cc}
    \toprule
    \multirow{2}{*}{{\tf{Model}}} & \multicolumn{2}{c}{\tf{ACE05}} & \multicolumn{2}{c}{\tf{SciERC}} \\
     & Rel & Speed & Rel & Speed\\
     & (F1) & (sent/s) & (F1) & (sent/s) \\
    \midrule
    Full (single) & \tf{66.7} & 32.1 & \tf{48.2} & 34.6 \\
    Approx. (single) & 65.7 & \tf{384.7} & 47.0 & \tf{301.1} \\
    \midrule
    Full (cross) & \tf{67.7} & 14.7 & \tf{50.1} & 19.9 \\
    Approx. (cross) & 66.5 & \tf{237.6} & 48.8 & \tf{194.7} \\
    \bottomrule
\end{tabular}
}
\vspace{-0.4em}
\caption{We compare our full relation model and the approximation model in both accuracy and speed. The accuracy is measured as the relation F1 (boundaries) on the test set. These results are obtained using BERT-base for ACE05 and SciBERT for SciERC in both single-sentence and cross-sentence settings. The speed is measured on a single NVIDIA GeForce 2080 Ti GPU with a batch size of $32$. }
\label{tab:approx}
\end{center}
\end{table}

% Entity types are not considered when we compute the F1 scores.

\subsection{Batch Computations and Speedup}
\label{sec:approx}

In Section~\ref{sec:method-approx}, we proposed an efficient approximation solution for the relation model, which enables us to re-use the computations of text tokens and batch multiple span pairs in one input sentence.
We evaluate this approximation model on ACE05 and SciERC.
Table~\ref{tab:approx} shows the relation F1 scores and the inference speed of the full relation model and the approximation model.
On both datasets, our approximation model significantly improves the efficiency of the inference process.\footnote{Note that we only applied this batch computation trick at inference time, because we observed that training with batch computation leads to a slightly (and consistently) worse result. We hypothesize that this is due to the impact of increased batch sizes. We still modified the position embedding and attention masks during training (without batching the instances though).}
For example, we obtain a $11.9\times$ speedup on ACE05 and a $8.7\times$ speedup on SciERC in the single-sentence setting.
By re-using a large part of computations, we are able to make predictions on the full ACE05 test set (2k sentences) in less than $10$ seconds on a single GPU. On the other hand, this approximation only leads to a small performance drop and the relaion F1 measure decreases by only $1.0\%$ and $1.2\%$ on ACE05 and SciERC respectively in the single-sentence setting. Considering the accuracy and efficiency of this approximation model, we expect it to be very effective to use in practice.

%!TEX root = main.tex

\section{Analysis}
\label{sec:analysis}

Despite its simple design and training paradigm, we have shown that our approach outperforms all previous joint models. In this section, we aim to take a deeper look and understand what contributes to its final performance.

\subsection{Importance of Typed Text Markers}
\label{sec:input-features}

Our key observation is that it is crucial to build different contextual representations for different pairs of spans and an early fusion of entity type information can further improve performance. To validate this, we experiment the following variants on both ACE05 and SciERC:

\vspace{0.5em}
\noindent\textbf{\textsc{Text}}: We use the span representations defined in the entity model (Section~\ref{sec:our-model}) and concatenate the hidden representations for the subject and the object, as well as their element-wise multiplication: $[\mathbf{h}_e(s_i), \mathbf{h}_e(s_j), \mathbf{h}_e(s_i) \odot \mathbf{h}_e(s_j)]$. This is similar to the relation model in \newcite{luan2018multi,luan2019general}.

\vspace{0.3em}
\noindent\textbf{\textsc{TextEType}}: We concatenate the span-pair representations from \textsc{Text} with entity type embeddings $\psi(e_i), \psi(e_j) \in \mathbb{R}^{d_E}$ ($d_E$ = 150).

\vspace{0.3em}
\noindent\textbf{\textsc{Markers}}: We use untyped entity types ($\langle\ts{S}\rangle$, $\langle\ts{/S}\rangle$, $\langle\ts{O}\rangle$, $\langle\ts{/O}\rangle$) at the input layer and concatenate the representations of two spans' starting points.

\vspace{0.3em}
\noindent\textbf{\textsc{MarkersEType}}: We concatenate the span-pair representations from \textsc{Markers} with entity type embeddings $\psi(e_i), \psi(e_j) \in \mathbb{R}^{d_E}$ ($d_E$ = 150).

\vspace{0.3em}
\noindent\textbf{\textsc{MarkersELoss}}: We also consider a variant which uses untyped markers but add another FFNN to predict the entity types of subject and object through auxiliary losses. This is similar to how the entity information is used in multi-task learning~\cite{luan2019general,wadden2019entity}.

\vspace{0.3em}
\noindent\textbf{\textsc{TypedMarkers}}: This is our final model described in Section~\ref{sec:our-model} with typed entity markers.

%!TEX root = ../main.tex

% \begin{table}[t]
% \begin{center}
% \scalebox{0.9}{
% \begin{tabular}{lcc}
%     \toprule
%     \tf{Features} & \tf{ACE05} (gold) & \tf{SciERC}  (gold) \\
%     \midrule
%     \textsc{Text} & 67.6 & 61.7 \\
%     \textsc{TextEType} & 68.2 & 63.6 \\
%     \textsc{Markers} & 70.5 & 68.2 \\
%     \textsc{MarkersEType} & 71.9 & 68.9 \\
%     \textsc{MarkersELoss} & 70.7 & 68.0 \\
%     \textsc{TypedMarkers} & \tf{73.1} & \tf{69.1} \\
%     \bottomrule
% \end{tabular}
% }
% \end{center}
% \vspace{-0.4em}
% \caption{\danqi{Change this to predicted entities or keep both.} Relation F1 scores on the development set of ACE05 and SciERC with different input features (the \ti{gold} entities are given). The results are obtained using BERT-base for ACE05 and SciBERT for SciERC, \ti{without} cross-sentence context.
% % The scores are based on taking gold entities as inputs.
% % \danqi{Don't use ACE05-G and SciERC-G. Just emphasize that in the caption here gold entities are used here.}
% }
% \label{tab:input}
% \vspace{-0.4em}
% \end{table}

\begin{table}[t]
\begin{center}
\scalebox{0.9}{
\begin{tabular}{lcccc}
    \toprule
    \multirow{2}{*}{{\tf{Input}}} & \multicolumn{2}{c}{\tf{ACE05}} & \multicolumn{2}{c}{\tf{SciERC}} \\
    & gold & e2e & gold & e2e \\
    \midrule
    \textsc{Text} & 67.6 & 61.6 & 61.7 & 45.3 \\
    \textsc{TextEType} & 68.2 & 62.6 & 63.6 & 45.7 \\
    \textsc{Markers} & 70.5  & 63.3 & 68.2 & 49.1  \\
    \textsc{MarkersEType} & 71.3 & 63.8 & 68.9 & \tf{49.7} \\
    \textsc{MarkersELoss} & 70.7 & 63.6  & 68.0 & 49.2  \\
    \textsc{TypedMarkers} & \tf{72.6} & \tf{64.2} & \tf{69.1} & \tf{49.7} \\
    \bottomrule
\end{tabular}
}
\end{center}
\vspace{-0.4em}
\caption{Relation F1 (boundaries)  on the development set of ACE05 and SciERC with different input features.
\ti{e2e}: the entities are predicted by our entity model; \ti{gold}: the gold entities are given. The results are obtained using BERT-base with \ti{single-sentence} context for ACE05 and SciBERT with \ti{cross-sentence} context for SciERC. For both ACE05 and SciERC, we use the same entity models with \ti{cross-sentence} context to compute the \ti{e2e} scores of using different input features.
% We do not consider entity types when we compute the F1 scores.
% The scores are based on taking gold entities as inputs.
% \danqi{Don't use ACE05-G and SciERC-G. Just emphasize that in the caption here gold entities are used here.}
% \danqi{Are you confident about these numbers? They use the same entity model and the relation models are trained using gold data? Can you compile both dev and test numbers in one spreadsheet for all these variants? If you can confirm the numbers, we should include and discuss them. What makes me most confused is the entity F1 is much lower on SciERC but the gap between Markers and TypedMarkers is even smaller on ACE05. Do you have an explanation?}
% \zexuan{End-to-end numbers are very close. Should we even include them?}
% \zexuan{Oh. I just realized that some e2e numbers on ACE05 are based on an entity model with context window $>0$, because I used some numbers from our EMNLP submission (in which the entity model is with context). I will update the numbers soon.}
}
\label{tab:input}
\end{table}

Table~\ref{tab:input} summarizes the results of all the variants using either \ti{gold} entities or \ti{predicted} entities from the entity model. As is shown, different input representations make a clear difference and the variants of using marker tokens are significantly better than standard text representations and this suggests the importance of learning different representations with respect to different pairs of spans. Compared to \ts{Text}, \ts{TypedMarkers} improved the F1 scores dramatically by $+5.0\%$ and $+7.4\%$ absolute when gold entities are given. With the predicted entities, the improvement is reduced as expected while it remains large enough.
Finally, entity type is useful in improving the relation performance and an early fusion of entity information is particularly effective (\ts{TypedMarkers} vs \ts{MarkersEType} and \ts{MarkersELoss}). We also find that \ts{MarkersEType} to perform even better than \ts{MarkersEloss} which suggests that using entity types directly as features is better than using them to provide training signals through auxiliary losses.

\subsection{Modeling Entity-Relation Interactions}
%!TEX root = ../main.tex

\begin{table}[t]
\centering

\scalebox{0.95}{
% \begin{tabular}{l c c }
%     \toprule
%     {Encoder shared?} &  \xmark & \cmark \\
%     \midrule
%     Entity F1 & \tf{88.8} & 87.7 \\
%     Relation F1 & \tf{64.8} & 64.4 \\
%     \bottomrule
% \end{tabular}
\begin{tabular}{c c c }
    \toprule
    {Shared encoder?} &  Enity F1 & Relation F1 \\
    \midrule
    \xmark & \tf{88.8} & \tf{64.8} \\
    \cmark & 87.7 & 64.4 \\
    \bottomrule
\end{tabular}
}
\caption{Relation F1 (boundaries) scores when entity and relation encoders are shared and \ti{not} shared on the ACE05 development set. This result is obtained from BERT-base models with cross-sentence context. }
\label{tab:shared-encoder}
% \vspace{-1em}
\end{table}

% We do not consider entity types when we compute the relation F1.

% % \scalebox{0.95}{
% \begin{table}[t]
% \centering

% \scalebox{0.95}{
% \begin{tabular}{l c c c c}
%     \toprule
%     & \multicolumn{2}{c}{\tf{ACE05}} & \multicolumn{2}{c}{\tf{SciERC}} \\
%     Encoder shared? &  \xmark & \cmark & \xmark & \cmark \\
%     \midrule
%     Entity F1 & \tf{88.8} & 87.7 & \todo & \todo \\
%     Relation F1 & \tf{64.8} & 64.4 & \todo & \todo \\
%     \bottomrule
% \end{tabular}
% }
% \caption{We compare sharing and not sharing the entity and relation encoders on the development sets of ACE05 and SciERC. This result is obtained from BERT-base and SciBERT with cross-sentence context. }
% \label{tab:shared-encoder}
% \end{table}

% }

% \input{tables/error_propagation}
One main argument for joint models is that modeling the interactions between the two tasks can contribute to each other. In this section, we aim to validate if it is the case in our approach. We first study whether sharing the two representation encoders can improve performance or not. We train the entity and relation models together by jointly optimizing $\mathcal{L}_e + \mathcal{L}_r$ (Table~\ref{tab:shared-encoder}). We find that simply sharing the encoders hurts both the entity and relation F1. We think this is because the two tasks have different input formats and require different features for predicting entity types and relations, thus using separate encoders indeed learns better task-specific features. We also explore whether the relation information can improve the entity performance. To do so, we add an auxiliary loss to our entity model, which concatenates the two span representations as well as their element-wise multiplication (see the \textsc{Text} variant in Section~\ref{sec:input-features}) and predicts the relation type between the two spans ($r \in \mathcal{R}$ or $\epsilon$).  Through joint training with this auxiliary relation loss, we observe a negligible improvement ($<0.1\%$) on averaged entity F1 over $5$ runs on the ACE05 development set. To summarize, (1) entity information is clearly important in predicting relations (Section~\ref{sec:input-features}). However, we don't find that relation information to improve our entity model substantially\footnote{\newcite{miwa2016end} observed a  slight improvement on entity F1 by sharing the parameters (80.8 $\rightarrow$ 81.8 F1) on the ACE05 development data. \newcite{wadden2019entity} observed that their relation propagation layers improved the entity F1 slightly on SciERC but it hurts performance on ACE05.}; (2) simply sharing the encoders does not provide benefits to our approach.

% In the previous section, we have already shown that the entity information is useful in the relation model (either entity embeddings, auxiliary loss or input features) and the best way is through typed markers.

%!TEX root = ../main.tex

% \scalebox{0.95}{
\begin{table}[t]
\centering

\scalebox{0.92}{
\begin{tabular}{l c c}
    \toprule
     &  \tf{ACE05} & \tf{SciERC} \\
    \midrule
    Gold entities & \tf{64.8} & 49.7 \\
    \midrule
    10-way jackknifing & 63.9 & 48.1 \\
    \midrule
    $0.4n$ spans (typed) & 64.6 & \tf{50.2} \\
    $0.4n$ spans (untyped) & 56.9 & 48.4 \\
    $0.4n$ spans (untyped + eloss) & 63.0 & 48.5 \\
    \bottomrule
\end{tabular}
}
\vspace{-0.4em}
\caption{We compare relation F1 (boundaries) with different training strategies on the development sets of ACE05 and SciERC. This result is from training BERT-base and SciBERT models with cross-sentence context. \ti{typed}: typed markers, \ti{untyped}: untyped markers, \ti{untyped + eloss}: untyped markers with auxiliary entity loss. See text for more details.}
\label{tab:error-propagation}
% \vspace{-1em}
\end{table}

% }

 % We do not consider entity types when we compute the F1 scores.

\begin{figure}[t]
  \centering
  \includegraphics[scale=0.47]{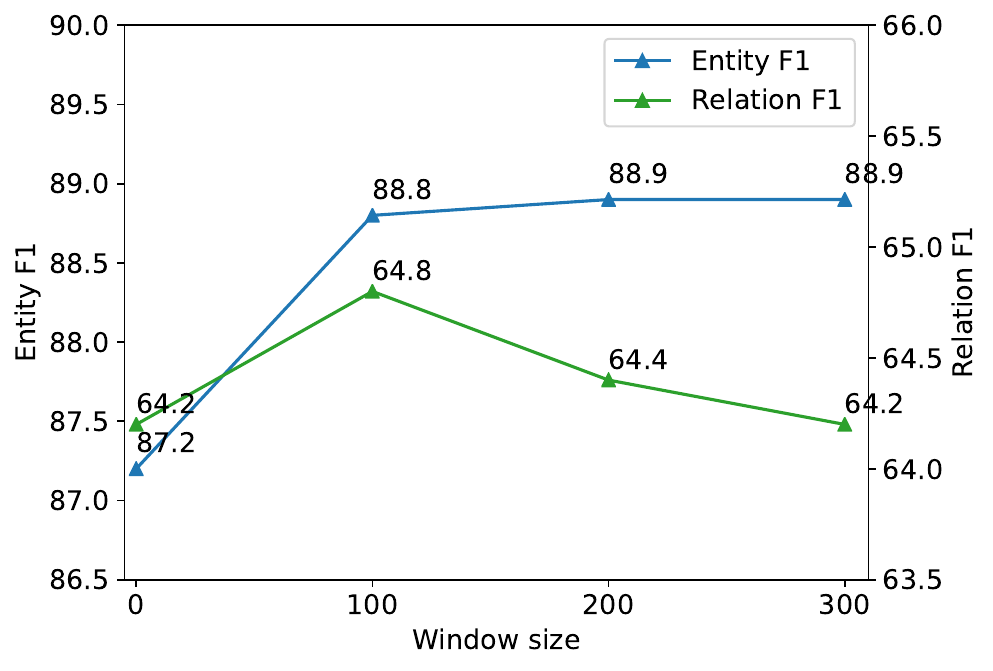}
    \vspace{-0.6em}
    \caption{Effect of different context window sizes, measured on the ACE05 development set with the BERT-base model. We use the same entity model (an entity model with $W=300$) to report the relation F1 scores (boundaries).}
    \label{fig:context_window}
    \vspace{-0.4em}
\end{figure}

\subsection{Mitigating Error Propagation}
\label{sec:error-prop}

A well-known drawback of pipeline training is the error propagation issue. In our final model, we use gold entities (and their types) to train the relation model and the predicted entities during inference and this may lead to a discrepancy between training and testing. In the following, we describe several attempts we made to address this issue.

We first study whether using predicted entities --- instead of gold entities --- during training can mitigate this issue. We adopt a 10-way jackknifing method, which is a standard technique in many NLP tasks such as dependency parsing~\cite{agic2017not}. Specifically, we divide the data into $10$ folds and predict the entities in the $k$-th fold using an entity model trained on the remainder. As shown in Table~\ref{tab:error-propagation}, we find that jackknifing strategy hurts the final relation performance surprisingly. We hypothesize that it is because it introduced additional noise during training.

Second, we consider using more pairs of spans for the relation model at both training and testing time.  The main reason is that in the current pipeline approach, if a gold entity is missed out by the entity model during inference, the relation model will not be able to predict any relations associated with that entity. Following the beam search strategy used in the previous work~\cite{luan2019general,wadden2019entity}, we consider using $\lambda n$ ($\lambda = 0.4$ and $n$ is the sentence length)\footnote{This pruning strategy achieves a recall of $96.7\%$ of gold relations on the development set of ACE05.} top spans scored by the entity model. We explored several different strategies for encoding the top-scoring spans for the relation model: (1) typed markers: the same as our main model except that we now have markers e.g., $\langle \textsc{S:}\epsilon\rangle$, $\langle \textsc{/S:}\epsilon\rangle$ as input tokens; (2) untyped markers: in this case, the relation model is unaware of a span is an entity or not; (3) untyped markers trained with an auxiliary entity loss ($e \in \mathcal{E}$ or $\epsilon$). As Table~\ref{tab:error-propagation} shows, none of these changes led to significant improvements and using untyped markers is especially worse because the relation model struggles to identify whether a span is an entity or not.

In sum, we do not find any of these attempts improved performance significantly and our simple pipelined training turns out to be a surprisingly effective strategy. We do not argue that this error propagation issue does not exist or cannot be solved, while we will need to explore better solutions to address this issue.

% A well-known drawback of pipelined training is the error propagation issue. In our final model, we use gold entities (and their types) to train the relation model and the predicted entities during inference and this may lead to a discrepancy between training and testing. We have explored several strategies to mitigate this possible error propagation issue, including 1) using predicted entities instead of gold entities through a 10-way jackknifing method; 2) considering more pairs of spans (we keep $0.4 n$ top scored spans using beam search, which achieves a recall of 96.7\% of gold relations on the development set of ACE05) for the relation model at both training and testing time. More details can be found in Appendix \ref{app:error-propagation}. As the results shown in Table~\ref{tab:error-propagation}, we don't find any of these attempts improved performance significantly and our simple pipelined training turns out to be a surprisingly effective strategy. We do not argue that this error propagation issue does not exist or cannot be solved, while we will need to explore better solutions to address this issue.

%

\subsection{Effect of Cross-sentence Context}
\label{sec:context}

In Table~\ref{tab:main-results}, we demonstrated the improvements from using cross-sentence context on both the entity and relation performance. We explore the effect of different context sizes $W$ in Figure~\ref{fig:context_window}. We find that using cross-sentence context clearly improves both entity and relation F1.
However, we find the relation performance doesn not further increase from $W = 100$ to $W = 300$.
In our final models, we use $W=300$ for the entity model and $W=100$ for the relation model.

%!TEX root = main.tex
\section{Conclusion}
\label{sec:conclusion}

In this paper, we present a simple and effective approach for end-to-end relation extraction. Our model learns two encoders for entity recognition and relation extraction independently and our experiments show that it outperforms previous state-of-the-art on three standard benchmarks considerably. We conduct extensive analyses to undertand the superior performance of our approach and validate the importance of learning distinct contextual representations for entities and relations and using entity information as input features for the relation model. We also propose an efficient approximation, obtaining a large speedup at inference time with a small reduction in accuracy. We hope that this simple model will serve as a very strong baseline and make us rethink the value of joint training in end-to-end relation extraction.

% \input{7_ethical}

% Acknowledgement: Yi Luan, Howard Chen, Karthik Narasimhan, Ameet Deshpande, Dan Friedman

\section*{Acknowledgements}
We thank Yi Luan for the help with the datasets and evaluation. We thank
Howard Chen, Ameet Deshpande, Dan Friedman, Karthik Narasimhan, and the anonymous reviewers for their helpful comments and feedback. This work is supported in part by a Graduate Fellowship at Princeton University.

\bibliography{refs}

\begin{thebibliography}{37}
\expandafter\ifx\csname natexlab\endcsname\relax\def\natexlab#1{#1}\fi

\bibitem[{Agi{\'c} and Schluter(2017)}]{agic2017not}
{\v{Z}}eljko Agi{\'c} and Natalie Schluter. 2017.
\newblock How (not) to train a dependency parser: The curious case of
  jackknifing part-of-speech taggers.
\newblock In \emph{Association for Computational Linguistics (ACL)}, pages
  679--684.

\bibitem[{Bekoulis et~al.(2018)Bekoulis, Deleu, Demeester, and
  Develder}]{bekoulis2018adversarial}
Giannis Bekoulis, Johannes Deleu, Thomas Demeester, and Chris Develder. 2018.
\newblock Adversarial training for multi-context joint entity and relation
  extraction.
\newblock In \emph{Empirical Methods in Natural Language Processing (EMNLP)},
  pages 2830--2836.

\bibitem[{Beltagy et~al.(2019)Beltagy, Lo, and Cohan}]{beltagy2019scibert}
Iz~Beltagy, Kyle Lo, and Arman Cohan. 2019.
\newblock Scibert: A pretrained language model for scientific text.
\newblock In \emph{Empirical Methods in Natural Language Processing (EMNLP)},
  pages 3606--3611.

\bibitem[{Bunescu and Mooney(2005)}]{bunescu2005shortest}
Razvan Bunescu and Raymond Mooney. 2005.
\newblock A shortest path dependency kernel for relation extraction.
\newblock In \emph{Empirical Methods in Natural Language Processing (EMNLP)},
  pages 724--731.

\bibitem[{Chan and Roth(2011)}]{chan2011exploiting}
Yee~Seng Chan and Dan Roth. 2011.
\newblock Exploiting syntactico-semantic structures for relation extraction.
\newblock In \emph{Association for Computational Linguistics: Human Language
  Technologies (ACL-HLT)}, pages 551--560.

\bibitem[{Devlin et~al.(2019)Devlin, Chang, Lee, and
  Toutanova}]{devlin2019bert}
Jacob Devlin, Ming-Wei Chang, Kenton Lee, and Kristina Toutanova. 2019.
\newblock {BERT}: Pre-training of deep bidirectional transformers for language
  understanding.
\newblock In \emph{North American Chapter of the Association for Computational
  Linguistics (NAACL)}, pages 4171--4186.

\bibitem[{Dixit and Al-Onaizan(2019)}]{dixit2019span}
Kalpit Dixit and Yaser Al-Onaizan. 2019.
\newblock Span-level model for relation extraction.
\newblock In \emph{Association for Computational Linguistics (ACL)}, pages
  5308--5314.

\bibitem[{Florian et~al.(2004)Florian, Hassan, Ittycheriah, Jing, Kambhatla,
  Luo, Nicolov, and Roukos}]{florian2004statistical}
Radu Florian, Hany Hassan, Abraham Ittycheriah, Hongyan Jing, Nanda Kambhatla,
  Xiaoqiang Luo, H~Nicolov, and Salim Roukos. 2004.
\newblock A statistical model for multilingual entity detection and tracking.
\newblock In \emph{North American Chapter of the Association for Computational
  Linguistics: Human Language Technologies (NAACL-HLT)}, pages 1--8.

\bibitem[{Florian et~al.(2006)Florian, Jing, Kambhatla, and
  Zitouni}]{florian2006factorizing}
Radu Florian, Hongyan Jing, Nanda Kambhatla, and Imed Zitouni. 2006.
\newblock Factorizing complex models: A case study in mention detection.
\newblock In \emph{Association for Computational Linguistics (ACL)}, pages
  473--480.

\bibitem[{Fu et~al.(2019)Fu, Li, and Ma}]{fu2019graphrel}
Tsu-Jui Fu, Peng-Hsuan Li, and Wei-Yun Ma. 2019.
\newblock {GraphRel}: Modeling text as relational graphs for joint entity and
  relation extraction.
\newblock In \emph{Association for Computational Linguistics (ACL)}, pages
  1409--1418.

\bibitem[{He et~al.(2018)He, Lee, Levy, and Zettlemoyer}]{he2018jointly}
Luheng He, Kenton Lee, Omer Levy, and Luke Zettlemoyer. 2018.
\newblock Jointly predicting predicates and arguments in neural semantic role
  labeling.
\newblock In \emph{Association for Computational Linguistics (ACL)}, pages
  364--369.

\bibitem[{Kambhatla(2004)}]{kambhatla2004combining}
Nanda Kambhatla. 2004.
\newblock Combining lexical, syntactic, and semantic features with maximum
  entropy models for information extraction.
\newblock In \emph{Association for Computational Linguistics (ACL)}, pages
  178--181.

\bibitem[{Katiyar and Cardie(2017)}]{katiyar2017going}
Arzoo Katiyar and Claire Cardie. 2017.
\newblock Going out on a limb: Joint extraction of entity mentions and
  relations without dependency trees.
\newblock In \emph{Association for Computational Linguistics (ACL)}, pages
  917--928.

\bibitem[{Lan et~al.(2020)Lan, Chen, Goodman, Gimpel, Sharma, and
  Soricut}]{lan2020albert}
Zhenzhong Lan, Mingda Chen, Sebastian Goodman, Kevin Gimpel, Piyush Sharma, and
  Radu Soricut. 2020.
\newblock {ALBERT}: A lite bert for self-supervised learning of language
  representations.
\newblock In \emph{International Conference on Learning Representations
  (ICLR)}.

\bibitem[{Lee et~al.(2017)Lee, He, Lewis, and Zettlemoyer}]{lee2017end}
Kenton Lee, Luheng He, Mike Lewis, and Luke Zettlemoyer. 2017.
\newblock End-to-end neural coreference resolution.
\newblock In \emph{Empirical Methods in Natural Language Processing (EMNLP)},
  pages 188--197.

\bibitem[{Li and Ji(2014)}]{li2014incremental}
Qi~Li and Heng Ji. 2014.
\newblock Incremental joint extraction of entity mentions and relations.
\newblock In \emph{Association for Computational Linguistics (ACL)}, pages
  402--412.

\bibitem[{Li et~al.(2019)Li, Yin, Sun, Li, Yuan, Chai, Zhou, and
  Li}]{li2019entity}
Xiaoya Li, Fan Yin, Zijun Sun, Xiayu Li, Arianna Yuan, Duo Chai, Mingxin Zhou,
  and Jiwei Li. 2019.
\newblock Entity-relation extraction as multi-turn question answering.
\newblock In \emph{Association for Computational Linguistics (ACL)}, pages
  1340--1350.

\bibitem[{Lin et~al.(2020)Lin, Ji, Huang, and Wu}]{lin2020joint}
Ying Lin, Heng Ji, Fei Huang, and Lingfei Wu. 2020.
\newblock A joint neural model for information extraction with global features.
\newblock In \emph{Association for Computational Linguistics (ACL)}.

\bibitem[{Luan et~al.(2018)Luan, He, Ostendorf, and Hajishirzi}]{luan2018multi}
Yi~Luan, Luheng He, Mari Ostendorf, and Hannaneh Hajishirzi. 2018.
\newblock Multi-task identification of entities, relations, and coreference for
  scientific knowledge graph construction.
\newblock In \emph{Empirical Methods in Natural Language Processing (EMNLP)},
  pages 3219--3232.

\bibitem[{Luan et~al.(2019)Luan, Wadden, He, Shah, Ostendorf, and
  Hajishirzi}]{luan2019general}
Yi~Luan, Dave Wadden, Luheng He, Amy Shah, Mari Ostendorf, and Hannaneh
  Hajishirzi. 2019.
\newblock A general framework for information extraction using dynamic span
  graphs.
\newblock In \emph{North American Chapter of the Association for Computational
  Linguistics (NAACL)}, pages 3036--3046.

\bibitem[{Miwa and Bansal(2016)}]{miwa2016end}
Makoto Miwa and Mohit Bansal. 2016.
\newblock End-to-end relation extraction using {LSTMs} on sequences and tree
  structures.
\newblock In \emph{Association for Computational Linguistics (ACL)}, pages
  1105--1116.

\bibitem[{Miwa and Sasaki(2014)}]{miwa2014modeling}
Makoto Miwa and Yutaka Sasaki. 2014.
\newblock Modeling joint entity and relation extraction with table
  representation.
\newblock In \emph{Empirical Methods in Natural Language Processing (EMNLP)},
  pages 1858--1869.

\bibitem[{Peters et~al.(2019)Peters, Neumann, Logan, Schwartz, Joshi, Singh,
  and Smith}]{peters2019knowledge}
Matthew~E Peters, Mark Neumann, Robert Logan, Roy Schwartz, Vidur Joshi, Sameer
  Singh, and Noah~A Smith. 2019.
\newblock Knowledge enhanced contextual word representations.
\newblock In \emph{Empirical Methods in Natural Language Processing (EMNLP)},
  pages 43--54.

\bibitem[{Ratinov and Roth(2009)}]{ratinov2009design}
Lev Ratinov and Dan Roth. 2009.
\newblock Design challenges and misconceptions in named entity recognition.
\newblock In \emph{Computational Natural Language Learning (CoNLL)}, pages
  147--155.

\bibitem[{Sang and De~Meulder(2003)}]{sang2003introduction}
Erik Tjong~Kim Sang and Fien De~Meulder. 2003.
\newblock Introduction to the conll-2003 shared task: Language-independent
  named entity recognition.
\newblock In \emph{Computational Natural Language Learning (CoNLL)}, pages
  142--147.

\bibitem[{Soares et~al.(2019)Soares, FitzGerald, Ling, and
  Kwiatkowski}]{soares2019matching}
Livio~Baldini Soares, Nicholas FitzGerald, Jeffrey Ling, and Tom Kwiatkowski.
  2019.
\newblock Matching the blanks: Distributional similarity for relation learning.
\newblock In \emph{Association for Computational Linguistics (ACL)}, pages
  2895--2905.

\bibitem[{Sun et~al.(2019)Sun, Gong, Wu, Gong, Jiang, Lan, Sun, and
  Duan}]{sun2019joint}
Changzhi Sun, Yeyun Gong, Yuanbin Wu, Ming Gong, Daxin Jiang, Man Lan, Shiliang
  Sun, and Nan Duan. 2019.
\newblock Joint type inference on entities and relations via graph
  convolutional networks.
\newblock In \emph{Association for Computational Linguistics (ACL)}, pages
  1361--1370.

\bibitem[{Taill{\'e} et~al.(2020)Taill{\'e}, Guigue, Scoutheeten, and
  Gallinari}]{taille2020sincere}
Bruno Taill{\'e}, Vincent Guigue, Geoffrey Scoutheeten, and Patrick Gallinari.
  2020.
\newblock Let's stop incorrect comparisons in end-to-end relation extraction!
\newblock In \emph{Empirical Methods in Natural Language Processing (EMNLP)}.

\bibitem[{Wadden et~al.(2019)Wadden, Wennberg, Luan, and
  Hajishirzi}]{wadden2019entity}
David Wadden, Ulme Wennberg, Yi~Luan, and Hannaneh Hajishirzi. 2019.
\newblock Entity, relation, and event extraction with contextualized span
  representations.
\newblock In \emph{Empirical Methods in Natural Language Processing (EMNLP)},
  pages 5788--5793.

\bibitem[{Wang and Lu(2020)}]{wang2020two}
Jue Wang and Wei Lu. 2020.
\newblock Two are better than one: Joint entity and relation extraction with
  table-sequence encoders.
\newblock In \emph{Empirical Methods in Natural Language Processing (EMNLP)}.

\bibitem[{Wolf et~al.(2019)Wolf, Debut, Sanh, Chaumond, Delangue, Moi, Cistac,
  Rault, Louf, Funtowicz, Davison, Shleifer, von Platen, Ma, Jernite, Plu, Xu,
  Scao, Gugger, Drame, Lhoest, and Rush}]{Wolf2019HuggingFacesTS}
Thomas Wolf, Lysandre Debut, Victor Sanh, Julien Chaumond, Clement Delangue,
  Anthony Moi, Pierric Cistac, Tim Rault, Rémi Louf, Morgan Funtowicz, Joe
  Davison, Sam Shleifer, Patrick von Platen, Clara Ma, Yacine Jernite, Julien
  Plu, Canwen Xu, Teven~Le Scao, Sylvain Gugger, Mariama Drame, Quentin Lhoest,
  and Alexander~M. Rush. 2019.
\newblock Huggingface's transformers: State-of-the-art natural language
  processing.
\newblock \emph{ArXiv}, abs/1910.03771.

\bibitem[{Zelenko et~al.(2002)Zelenko, Aone, and
  Richardella}]{zelenko2002kernel}
Dmitry Zelenko, Chinatsu Aone, and Anthony Richardella. 2002.
\newblock Kernel methods for relation extraction.
\newblock In \emph{Empirical Methods in Natural Language Processing (EMNLP)},
  pages 71--78.

\bibitem[{Zhang et~al.(2017{\natexlab{a}})Zhang, Zhang, and Fu}]{zhang2017end}
Meishan Zhang, Yue Zhang, and Guohong Fu. 2017{\natexlab{a}}.
\newblock End-to-end neural relation extraction with global optimization.
\newblock In \emph{Empirical Methods in Natural Language Processing (EMNLP)},
  pages 1730--1740.

\bibitem[{Zhang et~al.(2017{\natexlab{b}})Zhang, Zhong, Chen, Angeli, and
  Manning}]{zhang2017tacred}
Yuhao Zhang, Victor Zhong, Danqi Chen, Gabor Angeli, and Christopher~D.
  Manning. 2017{\natexlab{b}}.
\newblock Position-aware attention and supervised data improve slot filling.
\newblock In \emph{Empirical Methods in Natural Language Processing (EMNLP)},
  pages 35--45.

\bibitem[{Zhang et~al.(2019)Zhang, Han, Liu, Jiang, Sun, and
  Liu}]{zhang2019ernie}
Zhengyan Zhang, Xu~Han, Zhiyuan Liu, Xin Jiang, Maosong Sun, and Qun Liu. 2019.
\newblock {ERNIE}: Enhanced language representation with informative entities.
\newblock In \emph{Association for Computational Linguistics (ACL)}, pages
  1441--1451.

\bibitem[{Zheng et~al.(2017)Zheng, Wang, Bao, Hao, Zhou, and
  Xu}]{zheng2017joint}
Suncong Zheng, Feng Wang, Hongyun Bao, Yuexing Hao, Peng Zhou, and Bo~Xu. 2017.
\newblock Joint extraction of entities and relations based on a novel tagging
  scheme.
\newblock In \emph{Association for Computational Linguistics (ACL)}, pages
  1227--1236.

\bibitem[{Zhou et~al.(2005)Zhou, Su, Zhang, and Zhang}]{zhou2005exploring}
GuoDong Zhou, Jian Su, Jie Zhang, and Min Zhang. 2005.
\newblock Exploring various knowledge in relation extraction.
\newblock In \emph{Association for Computational Linguistics (ACL)}, pages
  427--434.

\end{thebibliography}
\bibliographystyle{acl_natbib}

\clearpage

%!TEX root = main.tex
\appendix

\section{Datasets}
\label{app:datasets}

We use ACE04, ACE05, and SciERC datasets in our experiments. Table~\ref{tab:data_stats} shows the data statistics of each dataset.

The ACE04 and ACE05 datasets are collected from a variety of domains, such as newswire and online forums. We follow \newcite{luan2019general}'s preprocessing steps\footnote{We use the script provided by \newcite{luan2019general}: \url{https://github.com/luanyi/DyGIE/tree/master/preprocessing}.} and split ACE04 into 5 folds and ACE05 into train, development, and test sets.

The SciERC dataset is collected from 12 AI conference/workshop proceedings in four AI communities~\cite{luan2018multi}.
SciERC includes annotations for scientific entities, their relations, and coreference clusters. We ignore the coreference annotations in our experiments.
We use the processed dataset which is downloaded from the project website\footnote{\url{http://nlp.cs.washington.edu/sciIE/}} of \newcite{luan2018multi}.

\section{Implementation Details}
\label{app:impl-details}
We implement our models based on HuggingFace's \ti{Transformers} library~\cite{Wolf2019HuggingFacesTS}.
For the entity model, we follow \newcite{wadden2019entity} and set the width embedding size as $d_F = 150$ and use a 2-layer FFNN with $150$ hidden units and ReLU activations to predict the probability distribution of entity types:
\begin{align*}
    P_e(e \mid s_i) = \mathrm{softmax}(\mathbf{W}_e \text{FFNN}(\mathbf{h}_e(s_i)).
\end{align*}
For the relation model, we use a linear classifier on top of the span pair representation to predict the probability distribution of relation types:
\begin{align*}
    P_r(r|s_i, s_j) = \mathrm{softmax}(\mathbf{W}_r \mathbf{h}_r(s_i, s_j)).
\end{align*}
For our approximation model (Section~\ref{sec:approx}), we batch candidate pairs by adding $4$ markers for each pair to the end of the sentence, until the total number of tokens exceeds $250$.
We train our models with Adam optimizer of a linear scheduler with a warmup ratio of $0.1$. For all the experiments, we train the entity model for $100$ epochs, and a learning rate of 1e-5 for weights in pre-trained LMs, 5e-4 for others and a batch size of 16. We train the relation model for $10$ epochs with a learning rate of 2e-5 and a batch size of 32.

%!TEX root = ../main.tex

\begin{table}[t]
\centering

\scalebox{0.95}{
\begin{tabular}{lcccc}
    \toprule
    \multirow{2}{*}{{\tf{Training data}}} & \multicolumn{2}{c}{\tf{Ours}} & \multicolumn{2}{c}{\tf{Joint}} \\
    & Ent & Rel & Ent & Rel \\
    \midrule
    $10\%$ & \tf{82.0} & \tf{46.9} & 81.5 & 37.0 \\
    $25\%$ & \tf{84.9} & \tf{57.6} & 84.6 & 49.0 \\
    $50\%$ & 85.5 & \tf{61.9} & \tf{86.2} & 57.7 \\
    $100\%$ & 87.2 & \tf{63.4} & \tf{87.4} & 61.0 \\
    \bottomrule
\end{tabular}
}
\vspace{-0.4em}
\caption{F1 scores on ACE05 development set when only a subset of training samples ($10\%$, $25\%$, $50\%$, or $100\%$) are provided. }
\label{tab:data-scarce}
% \vspace{-1em}
\end{table}

\section{Performance with Varying Data Sizes}
\label{app:perf-data-sizes}
We compare our pipeline model to a joint model with 10\%, 25\%, 50\%, 100\% of training data on the ACE05 dataset. Here, our goal is to understand whether our finding still holds when the training data is smaller (and hence it is expected to have more errors in entity predictions).

Our baseline of joint model is our reimplementation of DYGIE++~\cite{wadden2019entity},  without using propagation layers (the encoders are shared for the entity and relation model and no input marker is used; the top scoring $0.4n$ entities are considered in beam pruning).
As shown in Table~\ref{tab:data-scarce}, we find that our model achieves even larger gains in relation F1 over the joint model, when the number of training examples is reduced.
This further highlights the importance of explicitly encoding entity boundaries and type features in data-scarce scenarios.

\end{document}